\documentclass[10pt, a4paper]{article}

\usepackage[final]{lrec-coling2024}
\usepackage{booktabs}
\usepackage{afterpage}
\usepackage{listings}
\usepackage{amsmath}
\usepackage{makecell}
\usepackage{multirow}
\title{Social Orientation: A New Feature for Dialogue Analysis}

\name{Todd Morrill$^{\ast}$, Zhaoyuan Deng$^{\ast}$, Yanda Chen$^{\ast}$, Amith Ananthram$^{\ast}$, \\
    {\bf \large Colin Wayne Leach$^{\dagger}$}, {\bf \large Kathleen McKeown$^{\ast}$}} 

\address{Columbia University$^{\ast}$ Barnard College$^{\dagger}$\\
\{tm3229, zd2286, amith.ananthram\}@columbia.edu \\
\{yanda.chen, kathy\}@cs.columbia.edu, cleach@barnard.edu}

\abstract{
    There are many settings where it is useful to predict and explain the success or failure of a dialogue. Circumplex theory from psychology models the social orientations (e.g., \textit{Warm-Agreeable}, \textit{Arrogant-Calculating}) of conversation participants and can be used to predict and explain the outcome of social interactions. Our work is novel in its systematic application of social orientation tags to modeling conversation outcomes. In this paper, we introduce a new data set of dialogue utterances machine-labeled with social orientation tags. We show that social orientation tags improve task performance, especially in low-resource settings, on both English and Chinese language benchmarks. We also demonstrate how social orientation tags help explain the outcomes of social interactions when used in neural models. Based on these results showing the utility of social orientation tags for dialogue outcome prediction tasks, we release our data sets, code, and models that are fine-tuned to predict social orientation tags on dialogue utterances.
 \\ \newline  }

\begin{document}

\maketitleabstract

\section{Introduction}
Predicting and explaining the outcome of a conversation is important in many real-life settings. Customer service interactions, business negotiations, and diplomatic discussions between governments are often contentious \cite{sun-etal-2022-tracking}. The maintainers of Wikipedia often engage in heated discussions about page edits requiring moderator oversight \cite{zhang-etal-2018-conversations, chang-danescu-niculescu-mizil-2019-trouble}. On the website Reddit, communities such as \verb|r/changemyview| provide a place for users to discuss controversial topics. We define success as any conversation that remains respectful and free from personal attacks and inappropriate content. In all of these settings, it is useful to predict whether a conversation will succeed or fail. It may be even more important to perform a post-hoc analysis of a conversation to understand why it ended the way that it did. This can help moderators quickly remediate issues \cite{10.1145/3555095, 10.1145/3555603} and also help diplomats and business negotiators understand why a conversation ended in a deal or a deadlock \cite{peskov-etal-2020-takes}. Our work demonstrates: 1) how to make better dialogue outcome predictions using a novel set of social orientation features from psychology, and 2) how to use these features to explain what led to a conversation success or failure.

\afterpage{
  \begin{figure*}[t!]
    \centering
    \includegraphics[width=\textwidth]{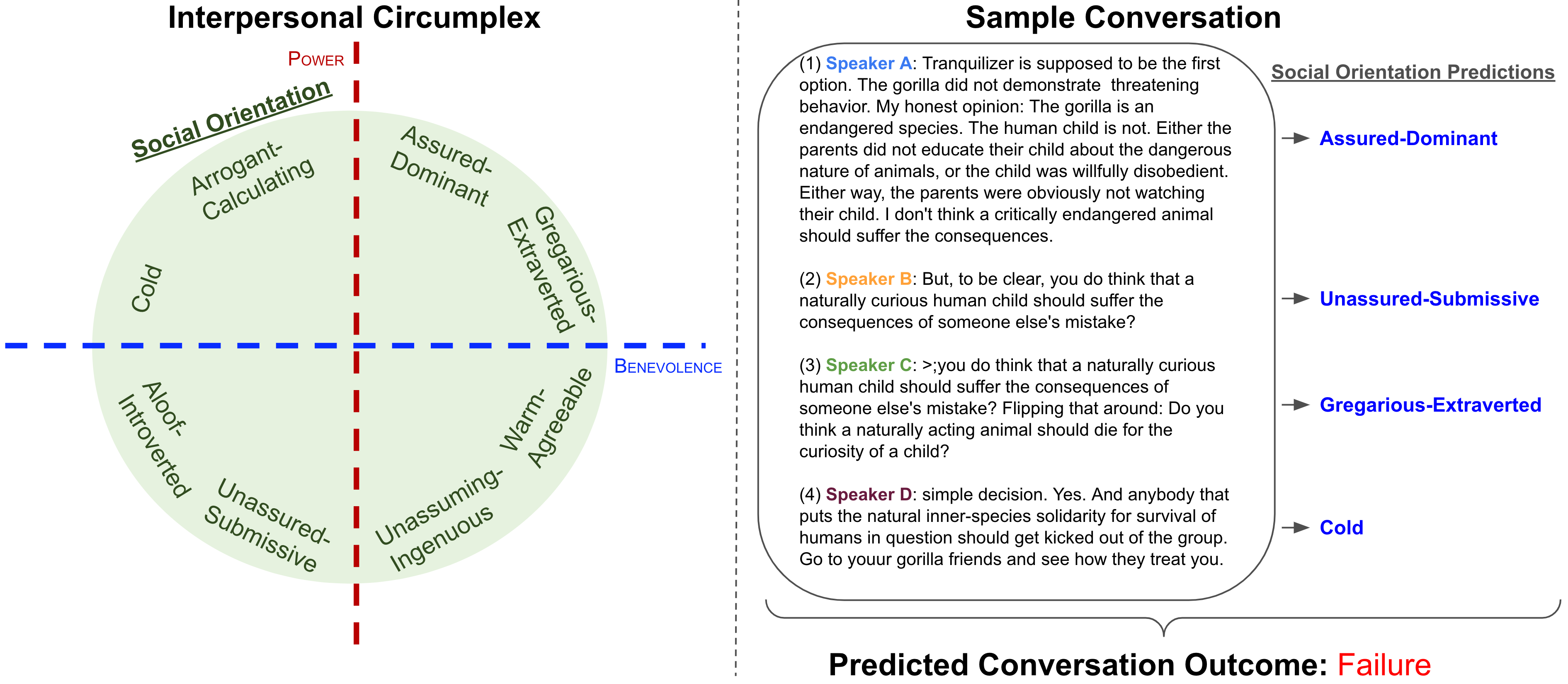}
    \caption{The left side of the figure shows the circumplex model for social orientations. The right side shows a sample conversation from the Reddit r/changemyview subreddit that ended in \textcolor{red}{failure}. Predicted social orientation tags are shown in \textcolor{blue}{blue}.}
    \label{fig:circumplex_sample_conversation}
  \end{figure*}
}    

Desirable properties of a dialogue modeling system include accuracy, generalizing from few labeled examples, low-cost of operation, fast inference, interpretability, and multi-lingual capabilities. It is common to start dialogue outcome prediction tasks with very few labeled conversation outcomes \cite{hu2022incontext}, which motivates the need for data efficient methods. Furthermore, while large language models such as GPT-4 \cite{openai2023gpt4} are effective at dialogue analysis, they are expensive to operate at scale \cite{10.1145/3442188.3445922, chen2023frugalgpt}. Computational budgets can also be important. For instance, deploying to laptops or other mobile devices makes it challenging to use large deep learning models \cite{sanh2020distilbert}. Finally, it is often valuable to support many languages and transfer linguistic resources from one language to another \cite{conneau-etal-2020-unsupervised}.

Circumplex theory and its social orientation tags can serve as a solution to the above concerns \cite{STRUS201770}. Circumplex theory posits that social interactions can be characterized by 2 dimensions - power and benevolence, where power refers to the degree to which an individual seeks to control, lead, or assert themselves in interpersonal relations, while benevolence measures the warmth, friendliness, and positivity of interactions. The combination of these two axes can describe social orientations that people may assume in social interactions, which are shown in Figure \ref{fig:circumplex_sample_conversation}. This theory can help explain which interaction styles are likely to end in success or failure, which may help moderators and participants in a conversation. For example, a conversation between two speakers both using \textit{Arrogant-Calculating} language is more likely to end in a deadlock than a conversation between a speaker using \textit{Arrogant-Calculating} language and a speaker using \textit{Unassured-Submissive} language as we show in our experiments below.

In this paper, we show that neural models that use text and social orientation features achieve state-of-the-art performance and enable explainability of dialogue outcomes on two widely used English dialogue benchmarks \cite{zhang-etal-2018-conversations, chang-danescu-niculescu-mizil-2019-trouble} and a new Chinese dialogue benchmark \cite{hua-etal-2018-wikiconv}. We also show that simple logistic regression classifiers that rely only on social orientation features outperform large pre-trained transformers in low-resource settings. To our knowledge, this work is the first to use social orientation features for predicting and explaining dialogue outcomes.

Our main contributions are:
\begin{enumerate}
    \item We release a new data set of dialogue utterances labeled with social orientation tags and a distilled model trained to predict these tags\footnote{Coming soon}.
    \item We achieve state-of-the-art task performance on 2 English dialogue outcome prediction data sets through the use of social orientation features in high-resource settings (i.e., neural models plus large data sets).
    \item We construct a new Chinese dialogue outcome prediction data set and show that applying social orientation features increases task performance in a second language.
    \item We demonstrate that including social orientation features in neural models increases explainability for dialogue outcome prediction tasks.
    \item We show that in low-resource settings, social orientation features are more effective than text-only neural models.
\end{enumerate}

The rest of this paper is organized as follows. Section \ref{sec:relatedwork} covers related work. Section \ref{sec:social-labeling} discusses data collection. Section \ref{sec:methods} describes our methods for predicting dialogue outcomes. Section \ref{sec:experiments} describes our experiments. Section \ref{sec:results} presents our results. Section \ref{sec:conclusion} concludes and discusses future work.

\section{Related Work}
\label{sec:relatedwork}
There have been many attempts to use the utterances in a conversation to predict its outcome. Outcomes may include the success or failure in the civility of an online conversation \cite{zhang-etal-2018-conversations, chang-danescu-niculescu-mizil-2019-trouble}, a deal struck in a negotiation \cite{lewis-etal-2017-deal}, a debate outcome \cite{zhang-etal-2016-conversational}, or whether or not someone has been persuaded to donate to charity \cite{wang-etal-2019-persuasion}. Some approaches emphasize end-to-end learning where task performance is the priority while others aim to use speaker demographics (e.g., age, sex, etc.), social network features \cite{10.1145/3442381.3449861}, etc. to enable explainability. Our work aims to deliver explainability while improving on task performance.

The Interpersonal Circumplex (for a review, see \cite{STRUS201770}) is part of a larger family of two-dimensional models in psychology that characterize personality traits, emotion, and person representation \cite{leach2015groups}. The circumplex specifies eight social orientations that people can adopt toward other people, which vary in their Power (the vertical dimension in Figure \ref{fig:circumplex_sample_conversation}) and Benevolence (the horizontal dimension in Figure \ref{fig:circumplex_sample_conversation}). We provide a detailed definition for each of the eight social orientation tags in the Appendix. Although circumplex has been used in psychology to examine the interaction between parties \cite{markey2003complementarity, sadler2009wavelength}, these studies used the two dimensions rather than the eight social orientations as we do here. Vassen and colleagues used circumplex theory to classify single sentences in Dutch \cite{vaassen-daelemans-2011-automatic} and instant message conversations \cite{vaasen2012delearyous}. To the best of our knowledge, this work is the first to use the eight social orientations of circumplex theory for dialogue outcome prediction tasks. Other approaches from the social sciences have also been successfully applied to dialogue tasks including sentiment tagging \cite{10.1613/jair.1.12802} and emotion tagging \cite{poria2019emotion}.

\section{Collecting Social Orientation Tags}
\label{sec:social-labeling}
We construct a new data set of dialogue utterances labeled with social orientation tags using GPT-4. Our complete prompt and sample inputs and outputs are provided in the Appendix. 

\subsection{Conversations Gone Awry}
We collect social orientation tags for all 30,021 utterances in the Wikipedia portion of the Conversations Gone Awry (CGA) corpus \cite{zhang-etal-2018-conversations, chang-danescu-niculescu-mizil-2019-trouble}\footnote{\url{https://convokit.cornell.edu/documentation/awry.html}}. Throughout the paper, we refer to this corpus as CGA. The CGA corpus is a collection of 4,188 conversations ranging from 3 to 20 turns. Conversations consist of editors trying to reach consensus about what to include/exclude from a Wikipedia page. These dialogues were initially identified from a much larger collection of Wikipedia discussion pages using a machine learning classifier to flag candidate conversations that likely contained a toxic contribution. Conversations were then manually reviewed by humans to assess whether the conversation contains a comment containing a personal attack. For the purposes of dialogue outcome modeling, if a conversation contains a personal attack it is labeled as a failed conversation and labeled successful otherwise. This toxic comment is then removed and the entire dialogue context before the attack can be used to predict if the conversation will succeed or fail. The data set has a balanced number of successful and unsuccessful conversations, making accuracy a useful measure of task performance.

\subsection{Summary Statistics}
\begin{figure}[!ht]
    \begin{center}
    \includegraphics[scale=0.45]{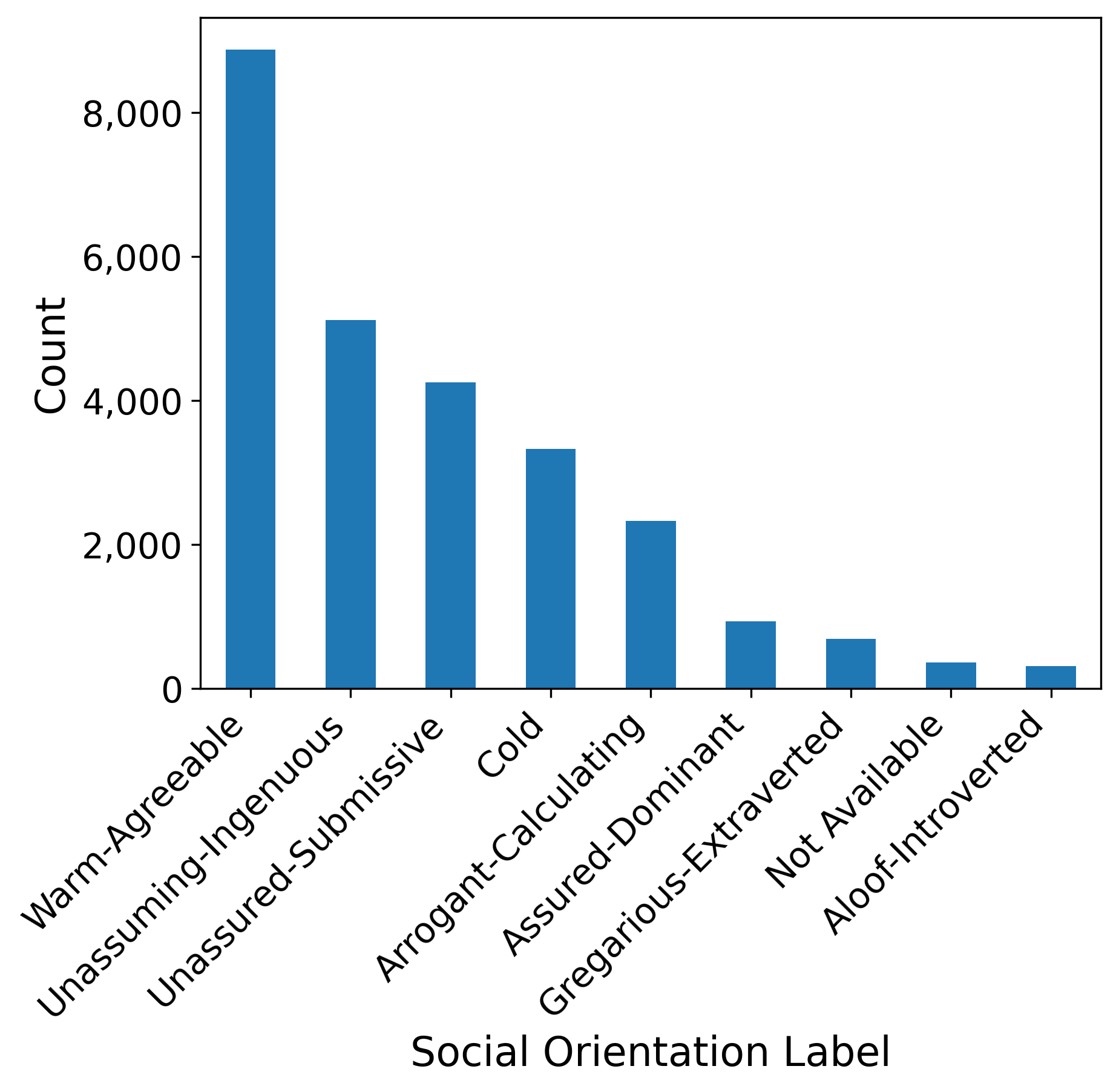} 
    \caption{Distribution of GPT-4 labeled social orientation tags in the Conversations Gone Awry (CGA) Corpus.}
    \label{fig:social-orientation-distribution}
    \end{center}
\end{figure}

The distribution of social orientation labels on the CGA data set from GPT-4 is given in Figure \ref{fig:social-orientation-distribution}. We note that while many utterances are tagged with \textit{Warm-Agreeable}, most other labels have support in the thousands, indicating that there is a wide range of social orientation tags represented in the data set.

We conduct a manual review of a sample of conversations from CGA to assess the degree to which humans agree on the social orientation tagging task as well as the agreement rate between humans and GPT-4. We evaluate this data collection procedure by measuring inter-annotator agreement among 3 annotators on a subset of 30 conversations each, with an overlap of 10 conversations. This results in 70 human labeled conversations for a total of 423 labeled dialogue utterances. Fleiss' Kappa for these 3 annotators is 0.42 indicating there is moderate agreement. To understand this number further, we compare the annotations between pairs of annotators and observe that the agreement rate ranges from 47\% to 59\%. Annotators were provided instructions, one round of feedback on a sample of their annotations, and then proceeded to complete their final annotations. We also compare human annotations to GPT-4 labels which ranges from 20\% to 30\% agreement. We found that most inter-annotator disagreements, including with GPT-4, occurred among neighboring tags (e.g., \textit{Cold} versus \textit{Arrogant-Calculating}). Humans also tended to use \textit{Assured-Dominant} more frequently than GPT-4. The complete set of metrics are reported in the Appendix. While GPT-4 has lower inter-annotator agreement with humans, its tags are nonetheless useful in downstream tasks, as our results show.

\section{Methods}
We use our \textit{GPT-4}-labeled utterances to train a distilled student social orientation tagger to allow us to efficiently label additional documents (Section~\ref{sec:social_model}). We evaluate the usefulness of both the original \textit{GPT-4} social orientation labels and the \textit{predicted} labels from our distilled model by using them as features in downstream dialogue outcome prediction tasks in English and Chinese (Section~\ref{sec:soc-orie-augm}).

\label{sec:methods}
\subsection{Social Orientation Prediction Model}
\label{sec:social_model}
In this work, we use social orientation tags as features in downstream dialogue outcome prediction models to assess their impact in low and high-resource settings and for explainability. After data collection, we distill GPT-4's labels into a student model using \verb|DistilBERT| \cite{sanh2020distilbert} for English and \verb|XLMR-base| \cite{conneau-etal-2020-unsupervised} for other languages, allowing the entire community to benefit from social orientation tags in a computationally and cost-efficient manner. We note that while the \verb|XLMR| model is trained on English language social orientation data, this multilingual model supports up to 100 other languages (e.g., Chinese) so it can be used make social orientation predictions in languages other than English.

\subsection{Social Orientation Augmented Conversation Outcome Prediction}
\label{sec:soc-orie-augm}
We evaluate the usefulness of different socio-linguistic attributes, including social orientation tags, by using them as features in models of increasing complexity on the task of dialogue outcome prediction in English and Chinese. This allows us to study both the discriminativeness of these features on their own and the ways they interact with neural text models.

Based on the complexity of the underlying model, these socio-linguistic features are represented as either count-based encodings of their frequency in an input document or are prepended as newly initialized trainable tokens to the text of the document before ingestion. For instance, a sample utterance with a social orientation tag prepended to the text would look like the following: \textit{SpeakerA (Warm-Agreeable): That sounds like a good plan.}

\section{Experiments}
\label{sec:experiments}
In this section we describe our experiments evaluating accuracy on dialogue outcome prediction tasks using a variety features and models. Section \ref{sec:eval-data-sets} describes the data sets used for evaluation. Section \ref{sec:dialogue-outcome-prediction} describes the socio-linguistic features and models used in the experiments.

\subsection{Evaluation Data Sets}
\label{sec:eval-data-sets}
\paragraph{CGA (en)}
We use a held-out test portion of the CGA corpus to assess the generalization of our methods to new unseen documents from the same domain as the training data.

\paragraph{CGA CMV (en)}
We use the Reddit Change My View (CMV) portion\footnote{\url{https://convokit.cornell.edu/documentation/awry_cmv.html}} of the CGA corpus as a held-out test data set to assess the generalization of our distilled social orientation tags to a new genre. Throughout the paper, we refer to this corpus as CGA CMV. The CGA CMV corpus is a collection of 6,842 Reddit conversations where a user posts a statement like, ``CMV: I believe that X is true'', other users try to change their view, and the conversation may or may not derail into personal attacks. We note that we have GPT-4 labeled social orientation tags for CGA (Wikipedia) and do not collect social orientation tags for CGA CMV (Reddit) to assess the generalization capabilities of our social orientation tagger with no additional GPT-4 costs. In practice, we use the \verb|DistilBERT| social orientation prediction model described in Section \ref{sec:social_model} to predict social orientation tags for utterances in the CGA CMV data set. We then train dialogue outcome prediction models using the training portion of the CGA CMV data set using these predicted social orientation tags as features.

\paragraph{WikiConv (zh)}
To assess the performance of our approach in a multilingual context, we construct a Chinese dialogue outcome prediction data set by adopting the methodology used for CGA \cite{zhang-etal-2018-conversations, chang-danescu-niculescu-mizil-2019-trouble}. Using COLDETECTOR \citep{deng-etal-2022-cold}, an off-the-shelf Chinese toxic utterance detector,  we identify potential toxic utterances within the Chinese portion of the WikiConv data set \citep{hua-etal-2018-wikiconv}. These utterances are then manually reviewed by two native Chinese speakers to confirm the presence of toxicity. We use all conversation turns before the first toxic comment to predict if the conversation will succeed or fail. Furthermore, to create a balanced data set, we pair each toxic conversation with a non-toxic one from the same Wikipedia talk page. A comprehensive description of the data set construction procedure is provided in the Appendix.
This yields a total of 468 paired successful and unsuccessful conversations, spanning 157 distinct talk pages. The average length of a conversation before the first toxic comment is 9.6 turns. We experiment with social orientation tags predicted using our \verb|XLMR| model described in Section~\ref{sec:social_model} as features in dialogue outcome prediction models on this data set.

\subsection{Conversation Outcome Prediction}
\label{sec:dialogue-outcome-prediction}
\paragraph{Socio-Linguistic Features}
We experiment with four features extracted from the dialogue:
\begin{enumerate}
\item \verb|TF-IDF| features \cite{sparckjones1972statistical}
\item Transformer embeddings: \verb|DistilBERT| for English and \verb|XLMR| for Chinese.
\item \verb|Sentiment| count: normalized counts of sentiment tags for dialogue utterances using a pre-trained sentiment tagger \cite{camacho-collados-etal-2022-tweetnlp}.
\item Social orientation tags: we experiment with \verb|GPT-4|-generated tags and \verb|Predicted| tags generated by the distilled models described in Section~\ref{sec:social_model}.
\end{enumerate}

\paragraph{Models}
We fine-tune two types of models for conversation outcome prediction:
\begin{enumerate}
\item \verb|Logistic|: a logistic regression model
\item Transformer-based models: \verb|DistilBERT| for English and \verb|XLMR| for Chinese.
\end{enumerate}

We experiment with different settings by pairing up \verb|Feature| with \verb|Model|, denoted as \verb|Model (Feature)|. We test the effectiveness of the social orientation tags by comparing the accuracy of models augmented with social orientation tags (\verb|DistilBERT - GPT-4|, \verb|DistilBERT - Predicted|, \verb|XLMR - Predicted|, \verb|Logistic - Predicted|) to the accuracy of a model not augmented with socio-linguistic features (\verb|DistilBERT|) or augmented with other socio-linguistic features. Note that \verb|Model - Source| denotes a model paired with a social orientation source.

\section{Results}
In this section we report results for our distilled social orientation tagger (Section \ref{sec:tagger}) and dialogue outcome prediction models (Sections \ref{sec:high-resource}). We demonstrate explainability of the model's predictions (Section \ref{sec:explain}) and present a qualitative analysis of social orientation features (Section \ref{sec:qual-analysis}).

\label{sec:results}
\subsection{Social Orientation Tagger}
\label{sec:tagger}
\begin{figure}[!ht]
    \begin{center}
    \includegraphics[scale=0.45]{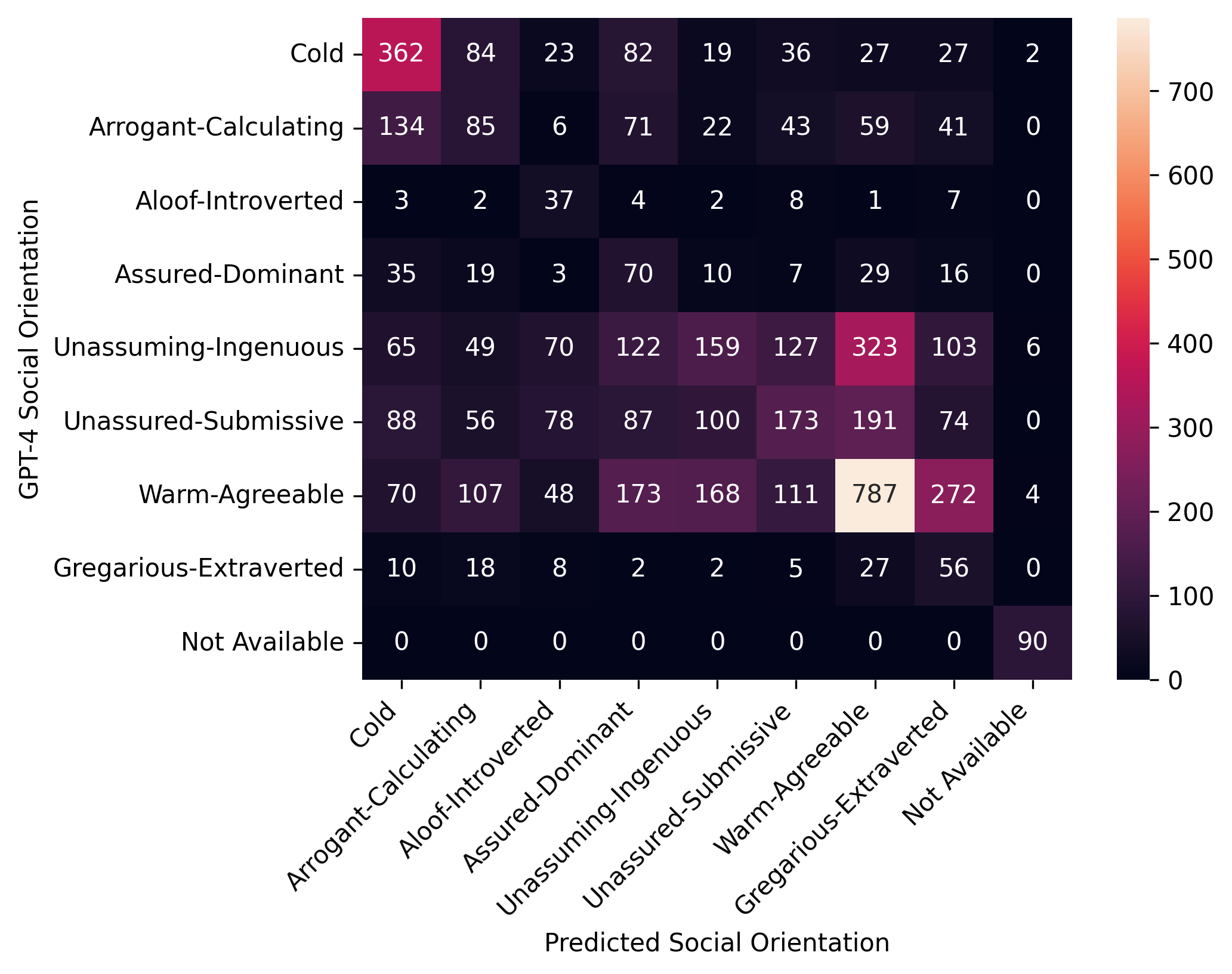} 
    \caption{Confusion matrix comparing predicted social orientation tags to GPT-4 labels on the CGA test set. We note that many of the misclassifications occur among neighboring tags.}
    \label{fig:social-confusion-matrix}
    \end{center}
\end{figure}

The student social orientation model achieves an accuracy of 35\% against the GPT-4 ground truth labels on the CGA corpus. Figure \ref{fig:social-confusion-matrix} shows that most disagreements tend to occur among neighboring tags (e.g., \textit{Unassuming-Ingenuous}, \textit{Unassured-Submissive}). In practice, this means that social orientation tags carry a lot of information. We use a weighted loss function, which reduces accuracy, but encourages more variety in the social orientation model's predictions, which we believe makes downstream results more explainable. Without loss weighting, it is possible to achieve an accuracy of 45\%.

\subsection{Social Orientation Tags Increase Accuracy}
\label{sec:high-resource}
 
\begin{figure}[!ht]
    \begin{center}
    \includegraphics[scale=0.35]{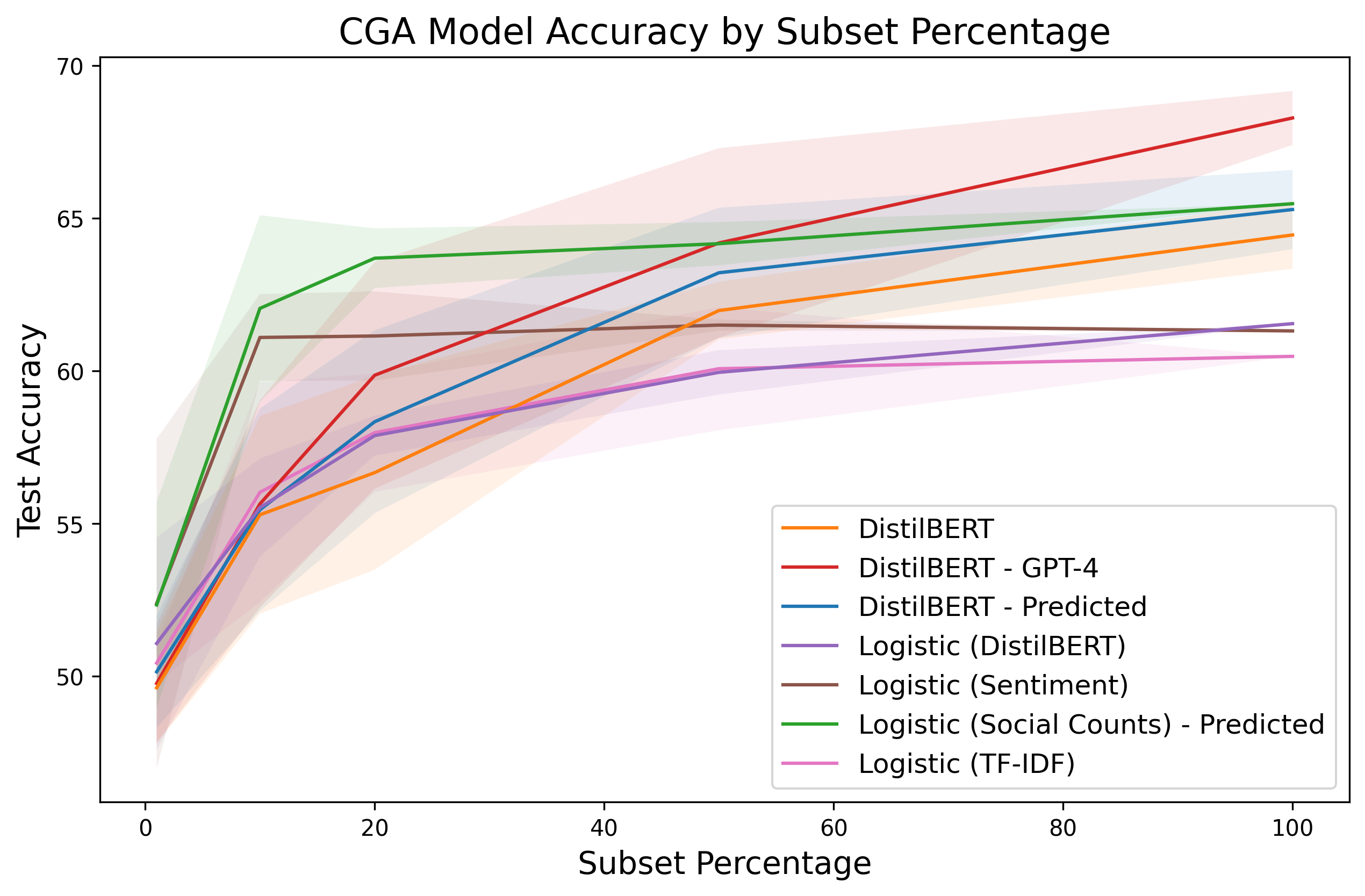}
    \includegraphics[scale=0.35]{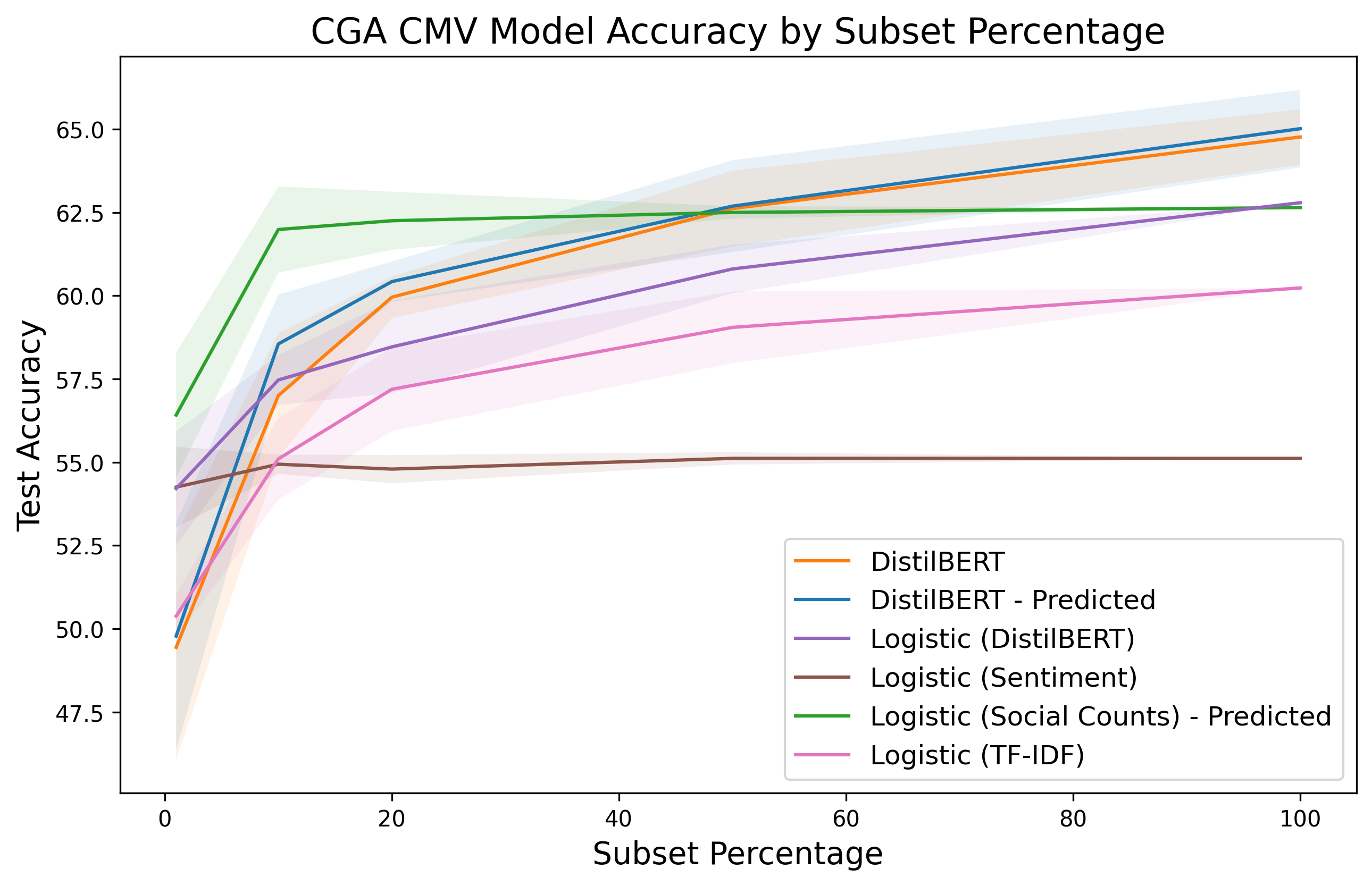}
    \caption{Data ablation results showing average accuracy scores ($\pm$ 1 standard deviation) over 5 runs for various methods on the CGA (top) and CGA CMV (bottom) data sets. We see that social orientation features (Logistic (Social Counts) - Predicted) outperform text-only methods in low-resource regimes and that social orientation features help model accuracy, even in high resource settings.}
    \label{fig:data-ablation}
    \end{center}
\end{figure}

\textbf{Social orientation features help deep learning models achieve state-of-the-art results.} We note that our best performing model on the CGA corpus is the \verb|DistilBERT - GPT-4| model that uses the text of the dialogue along with GPT-4 labeled social orientation tags. This result can be seen in Figure \ref{fig:data-ablation} (and in Table \ref{tab:combined-data-ablation}) in relation to all other experimental settings. This model achieves a 68.29\% accuracy, which to the best of our knowledge is a state-of-the-art (SOTA) result on this data set relative to the best known baseline of 66.50\% \cite{chang-danescu-niculescu-mizil-2019-trouble}. Figure \ref{fig:data-ablation} shows similar results on the CGA CMV data set. The \verb|DistilBERT - Predicted| model, which was trained on text and social orientation predictions from our distilled tagger, achieves an accuracy of 65.01\%, which we again believe to be SOTA relative to the best known baseline of 63.40\% \cite{chang-danescu-niculescu-mizil-2019-trouble}.

In Figure \ref{fig:data-ablation} and Table \ref{tab:combined-data-ablation} we see that the \verb|DistilBERT - Predicted| models trained on text plus predicted social orientation tags outperform text-only models (denoted \verb|DistilBERT|) on average at every data set size. This small but consistent improvement is an encouraging result.

\begin{figure}[!ht]
    \begin{center}
    \includegraphics[scale=0.35]{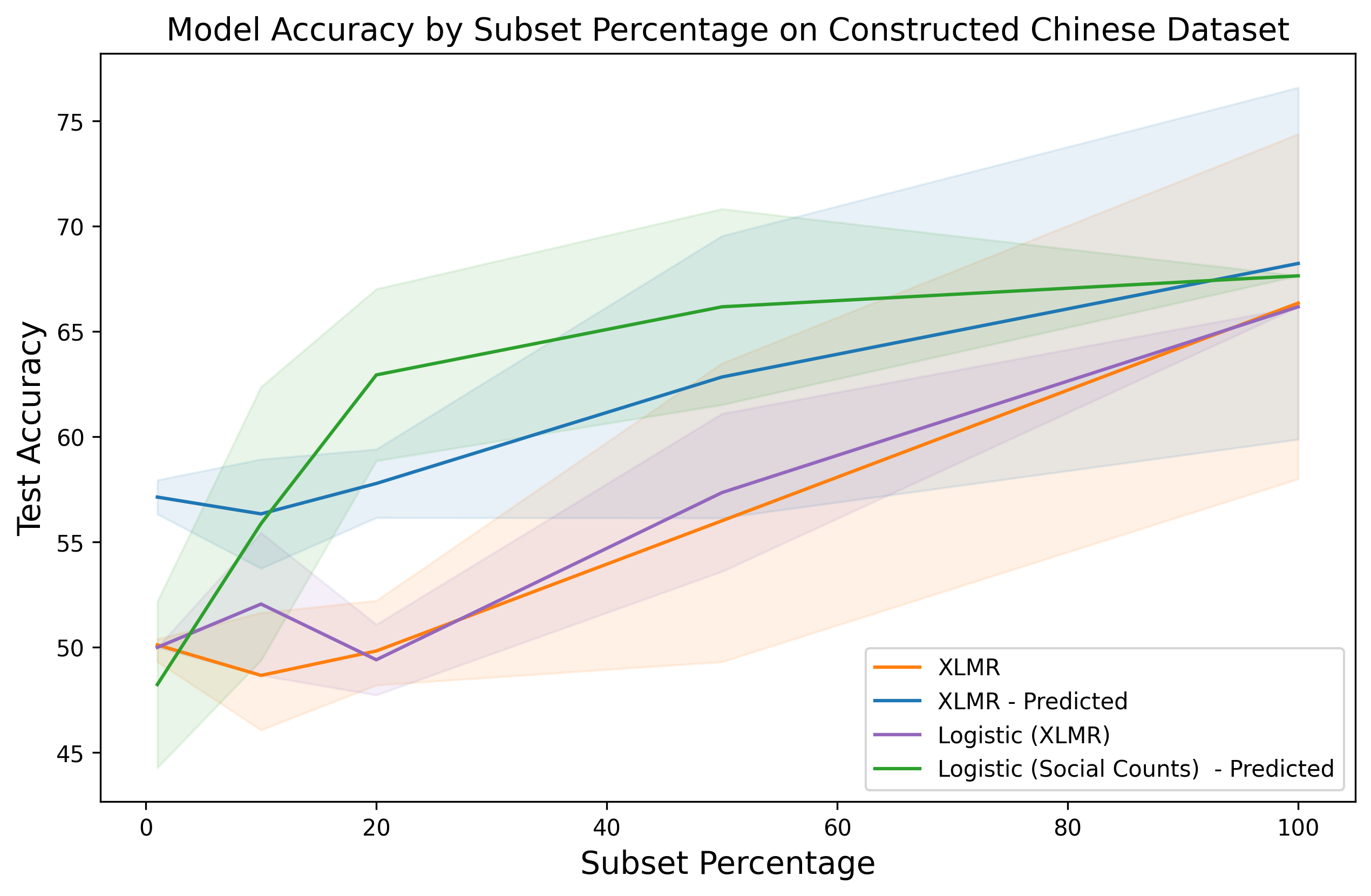}
    \caption{Data ablation results showing average accuracy scores ($\pm$ 1 standard deviation) over 5 runs for various methods on the constructed Chinese data set. We see high accuracy logistic regression results in low-resource settings and that text plus social orientation features outperform all other methods in high-resource settings.}
    \label{fig:cn-data-ablation}
    \end{center}
\end{figure}

Finally, Figure \ref{fig:cn-data-ablation} shows a similar result on the Chinese data set. The \verb|XLMR - Predicted| consistently outperforms the text-only \verb|XLMR| model at all data set sizes, confirming the effectiveness of social orientation tags in a second language.

\begin{table*}[t]

\centering
\begin{tabular}{clcllll}
\toprule
Datset&Subset \% & DistilBert & \makecell{DistilBert\\(Predicted)} & \makecell{ DistilBert\\(GPT-4)}&\makecell{Logistic\\(Predicted)}\\
\midrule
\multirow{5}{*}{CGA}&1 & 49.62 & 50.14 & 49.76 & \textbf{52.33\textsuperscript{*}}  \\
&10& 55.29 & 55.45 & 55.64 & \textbf{62.05\textsuperscript{*}} \\
&20 & 56.67 & 58.33 & 59.86\textsuperscript{*} & \textbf{63.69\textsuperscript{*}}\\
&50 & 61.98 & 63.21 &\textbf{64.19\textsuperscript{*}} & 64.17\textsuperscript{*}\\
&100 & 64.45 & 65.29 & \textbf{68.29\textsuperscript{*}} & 65.48\textsuperscript{*}\\
\midrule
\multirow{5}{*}{\makecell{CGA\\ CMV}}&1 & 49.44& 49.78 &-& \textbf{56.42\textsuperscript{*}} \\
&10 & 57.00& 58.55\textsuperscript{*}&-& \textbf{61.99\textsuperscript{*}} \\
&20 & 59.96 & 60.42&-& \textbf{62.25\textsuperscript{*}}  \\
&50 & 62.62 & \textbf{62.69}&- &62.50 \\
&100 & 64.77 & \textbf{65.01}& -&62.65  \\
\bottomrule
\end{tabular}
 \caption{Average accuracy scores comparing the performance of various models on CGA and CGA-CMV. We see that social orientation features outperform text features in low-resource regimes. * denotes t-test results where $p<0.1$ when compared to DistilBERT.}
 \label{tab:combined-data-ablation}
\end{table*}

\textbf{Social orientation features outperform text-only models in low-resource settings.}
\label{sec:low-resource}
Table \ref{tab:combined-data-ablation} shows test set accuracy scores averaged over 5 runs for the \verb|DistilBERT| model variants compared to the logistic regression model trained on normalized counts of social orientation tags in the conversation. Remarkably, this lightweight logistic regression model outperforms a 66 million parameter deep learning model in nearly all settings. We see similar results in the lower half of Table \ref{tab:combined-data-ablation}, which shows test set accuracy scores for the CGA CMV corpus. We see that with as few as about 400 conversations (0.1$\times$4,105), we're able to achieve over 60\% accuracy on this challenging dialogue outcome prediction task. The asterisks in Table \ref{tab:combined-data-ablation} indicate $p$-values at are significant at the level of $\leq0.1$. We see that the difference in performance between the logistic regression model and \verb|DistilBERT| text-only baseline is statistically significant up to about 20\% of the training data set size. Figure \ref{fig:data-ablation} (lower half for CGA CMV) shows the same results in a more visual format, where we see that the logistic regression model that uses social orientation count features outperforms all other methods in the low-resource setting including other logistic regression models trained on different feature sets. This confirms that it is indeed the social orientation tags that are improving accuracy in the low-resource setting.

Figure \ref{fig:cn-data-ablation} shows a similar result on the Chinese data set. The \verb|Logistic (Social Counts) - Predicted| model outperforms all other models in low-resource settings (i.e., when a 10\% - 50\% subset is used).

In summary, we have shown that the English language \verb|DistilBERT| and the multi-lingual \verb|XLMR| social orientation taggers can be used in conjunction with a logistic regression model trained on as few as a few hundred labeled conversations to perform dialogue outcome prediction tasks.

\subsection{Explainability of Model's Predictions}
\label{sec:explain}
\textbf{We gain explainability of the model's predictions through the use of circumplex theory.} 
We establish that the dialogue outcome prediction model is indeed using social orientation features in accordance with the theory by perturbing the social orientation tags and showing that the model changes its predictions accordingly. We first predict dialogue outcomes on the test portion of the CGA CMV corpus. We then randomly perturb all social orientation tags fed to the model and measure how many of the predictions change \cite{10.1145/2939672.2939778, lanham2023measuring}. If the model were ignoring social orientation features, we would expect very few prediction changes. Instead, we observe that the model changes its predictions on 275/1,368 conversations, or in other words, 20\% of the time, which shows that the model is indeed using social orientation tags to make its predictions. We note that the model still has access to the text of the dialogue, which carries a lot of information, so we wouldn't expect the model to change its predictions all of the time. 

\begin{figure}[!ht]
    \begin{center}
    \rule[0.2ex]{\linewidth}{1pt}
    \includegraphics[scale=0.29]{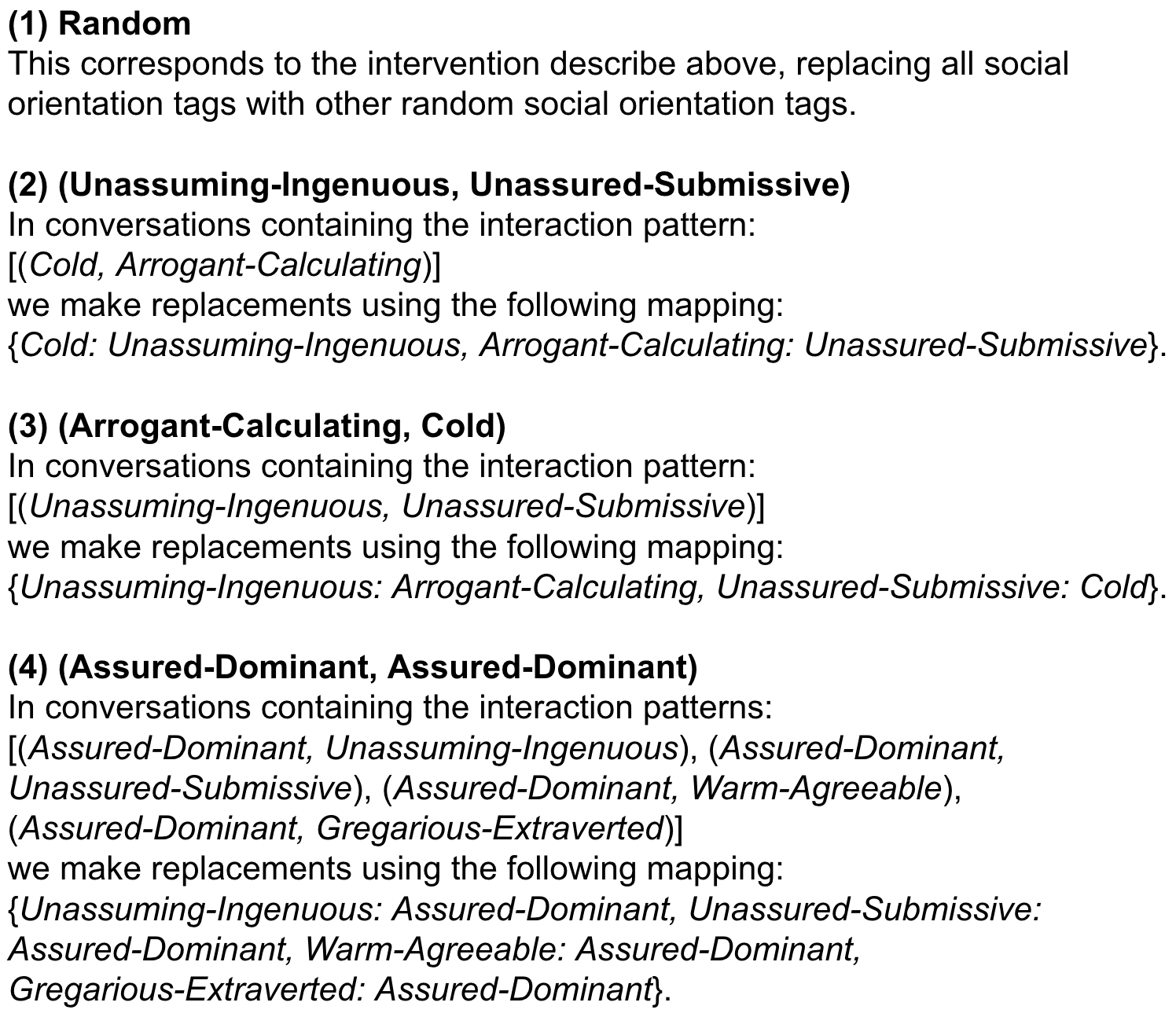} 
    \rule[0.2ex]{\linewidth}{1pt}
    \caption{Complete set of interventions for explainability experiments.}
    \label{fig:intervention}
    \end{center}
\end{figure}

Our complete set of interventions is enumerated in Figure \ref{fig:intervention}. Table \ref{tab:explain-interventions} shows complete results for our label perturbation experiments, where the Intervention column denotes what change was made, Pos2Neg denotes how many conversations were originally predicted to succeed but were predicted to fail after the intervention, Neg2Pos denotes the opposite of Pos2Neg, and Same denotes predictions that stayed the same.

\begin{table}[!ht]
    \begin{center}
    \resizebox{\columnwidth}{!}{
    \begin{tabular}{lrrr}
\toprule
Intervention & Pos2Neg & Neg2Pos & Same \\
\midrule
(1) Random & 81 & 194 & 1,093 \\
(2) (Unassuming-Ingenuous, Unassured-Submissive) & 0 & 154 & 333 \\
(3) (Arrogant-Calculating, Cold) & 15 & 0 & 64 \\
(4) (Assured-Dominant, Assured-Dominant) & 27 & 0 & 214 \\
\bottomrule
\end{tabular}

    }
    \end{center}
    \caption{Results showing the CGA CMV dialogue outcome prediction model changes its predictions in response to changes in the social orientation tags.}
    \label{tab:explain-interventions}
\end{table}

In order to further show that the model is making predictions in accordance with circumplex theory, we first state what circumplex would predict about an interaction pair and then show that the model behaves accordingly. For all such interventions, we first filter down to a subset of conversations in the CGA CMV test set and then replace social orientation tags according to the specification in Figure \ref{fig:intervention}. As an example, we describe intervention \textbf{(2) (Unassuming-Ingenuous, Unassured-Submissive)} here. We first identify all pairs of co-occurring social orientation tags within a conversation. We constrain co-occurring tags to be between \textit{different} speakers in the conversation because we are interested in the dynamics between speakers, not tags used by a single speaker. We then filter for conversations containing the interaction pattern (\textit{Cold}, \textit{Arrogant-Calculating}) or (\textit{Arrogant-Calculating}, \textit{Cold}), where we are ignoring order. Finally, we make the specified replacements, for instance in the case of intervention (2), we replace all \textit{Cold} tags with \textit{Unassuming-Ingenuous} and all \textit{Arrogant-Calculating} tags with \textit{Unassured-Submissive}. We then make a conversation outcome prediction for this modified conversation.

Our first targeted intervention is described under intervention (2) in Figure \ref{fig:intervention}. All else equal, circumplex theory would predict that conversations among participants using \textit{Unassuming-Ingenuous} and \textit{Unassured-Submissive} language are likely to succeed. After implementing the intervention, the model responds according to circumplex theory and we observe that 154/487 (31.6\%) conversations change their prediction from failure to success and 0 conversations change their prediction from success to failure.

We see a similar result for intervention (3) in the opposite direction. Circumplex theory would predict that the (\textit{Arrogant-Calculating}, \textit{Cold}) interaction pattern is more likely to fail. This is what we observe in the model's predictions after intervention (3), where 15/79 (19\%) conversations shift from predictions of success to predictions of failure and 0 shift their predictions from failure to success.

Our final targeted intervention, numbered (4), highlights the richness of this framework and distinguishes it from sentiment analysis. Sentiment tags often takes on the values \{\textit{negative}, \textit{neutral}, \textit{positive}\}. The \textit{Assured-Dominant} tag is a neutral tag in the sense that it is not overly negative or positive. If we solely relied on sentiment to predict whether the (\textit{Assured-Dominant}, \textit{Assured-Dominant}) interaction pattern is likely to succeed or fail, we would say that (\textit{neutral}, \textit{neutral}) is likely to succeed. In contrast, circumplex theory predicts that there is a likely chance of conflict when two or more conversations participants are using \textit{Assured-Dominant} language. Indeed, this is what we observe, where 27/241 (11.2\%) of conversations change their label from positive to negative and 0 conversations change their prediction from negative to positive.

\subsection{Qualitative Analysis}
\label{sec:qual-analysis}
We present 3 additional artifacts to demonstrate the utility of social orientation tags for dialogue outcome prediction tasks. First, we provide a sample conversation to show how social orientation tags can be used to explain the outcome of a conversation. Second, we show certain tags are more prevalent in conversations that derail, indicating their discriminative power. Finally, we show a co-occurrence matrix of social orientation tags to demonstrate that certain interaction patterns are more likely to end in success or failure.

\textbf{Social orientation tags help interpret conversations.} We refer back to Figure \ref{fig:circumplex_sample_conversation} to show how social orientation tags can be used to gain more interpretability of conversation outcomes. The conversation is a debate about whether the gorilla, Harambe, should have been killed after a 3-year old human boy fell into the gorilla's zoo enclosure. We see that the conversation starts with an utterance that asserts a number of statements about the situation in a ``firm'' manner which meets the definition of \textit{Assured-Dominant} (see Appendix for label definitions). This is followed by a comment that ``is not forceful,'' which meets the definition of \textit{Unassured-Submissive}. The next comment shows signs of someone who ``is not shy,'' since they ask a sequence of questions that encourages responses from both sides of the debate, which may meet the definition of \textit{Gregarious-Extraverted}. Finally, the conversation takes a turn on the last comment, which is predicted as \textit{Cold} (``is unsympathetic''). We see that the conversation starts out tense but civil and that the last speaker became more tactless, which may have caused the conversation to end in failure. This is an example of how social orientation tags can be used to interpret a conversation.

\begin{figure}[!ht]
    \begin{center}
    \includegraphics[scale=0.45]{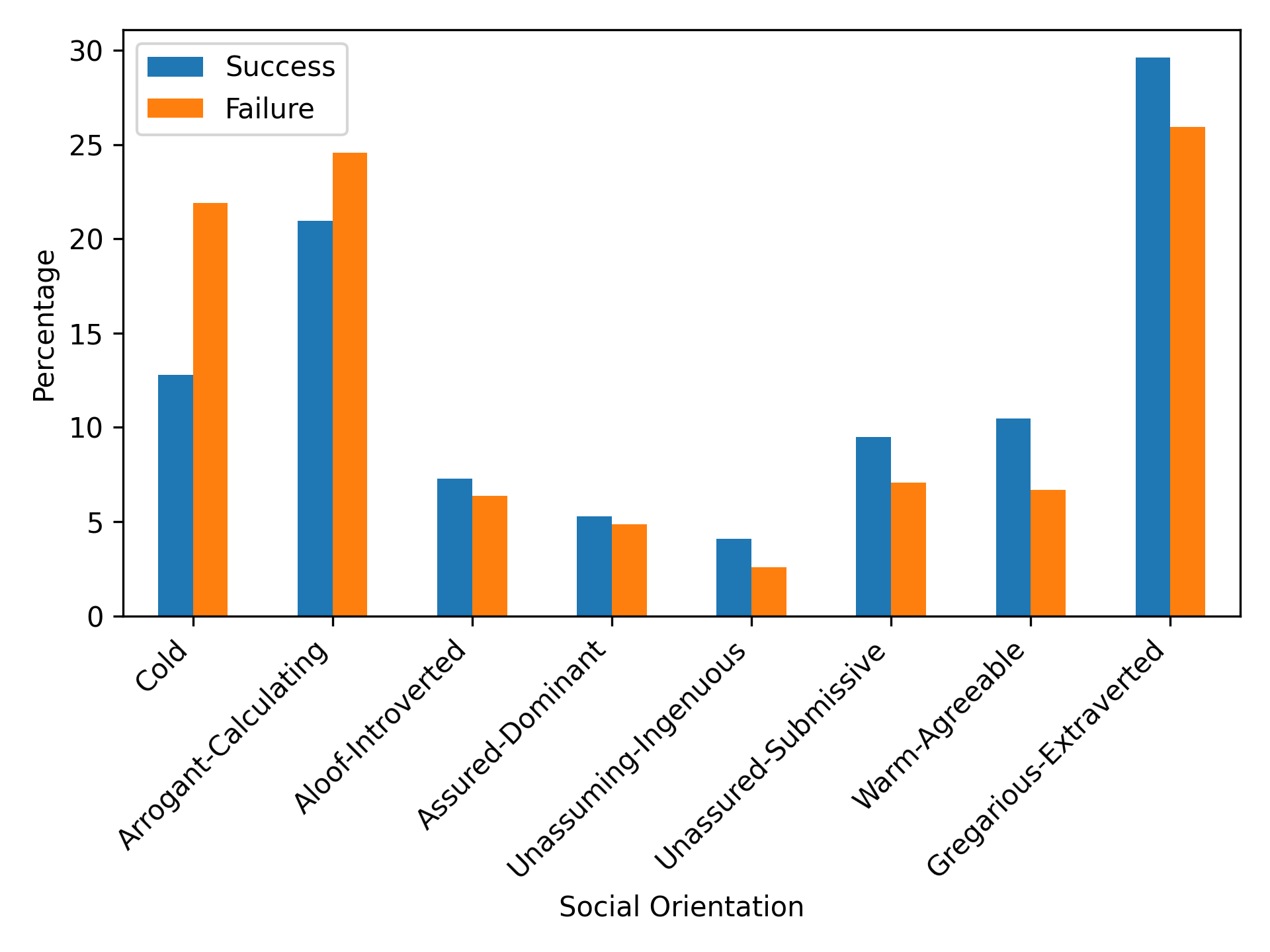}
    \caption{Social orientation tags by dialogue outcome for the CGA CMV corpus. Tags are generated by a DistilBERT model fine-tuned on the CGA corpus.}
    \label{fig:reddit-social-orientation-by-class}
    \end{center}
\end{figure}

\textbf{Social orientation features are discriminative with respect to dialogue outcomes.} Figure \ref{fig:reddit-social-orientation-by-class} shows the prevalence of social orientation tags by ground truth dialogue outcome in the CGA CMV data set. We see, for example, that \textit{Cold} and \textit{Arrogant-Calculating} tags are more prevalent in conversations that end in failure, while \textit{Warm-Agreeable} and \textit{Unassured-Submissive} tags are more prevalent in conversations that succeed. This indicates that these tags are useful for predicting dialogue outcomes.

\begin{figure}[!ht]
    \begin{center}
    \includegraphics[scale=0.45]{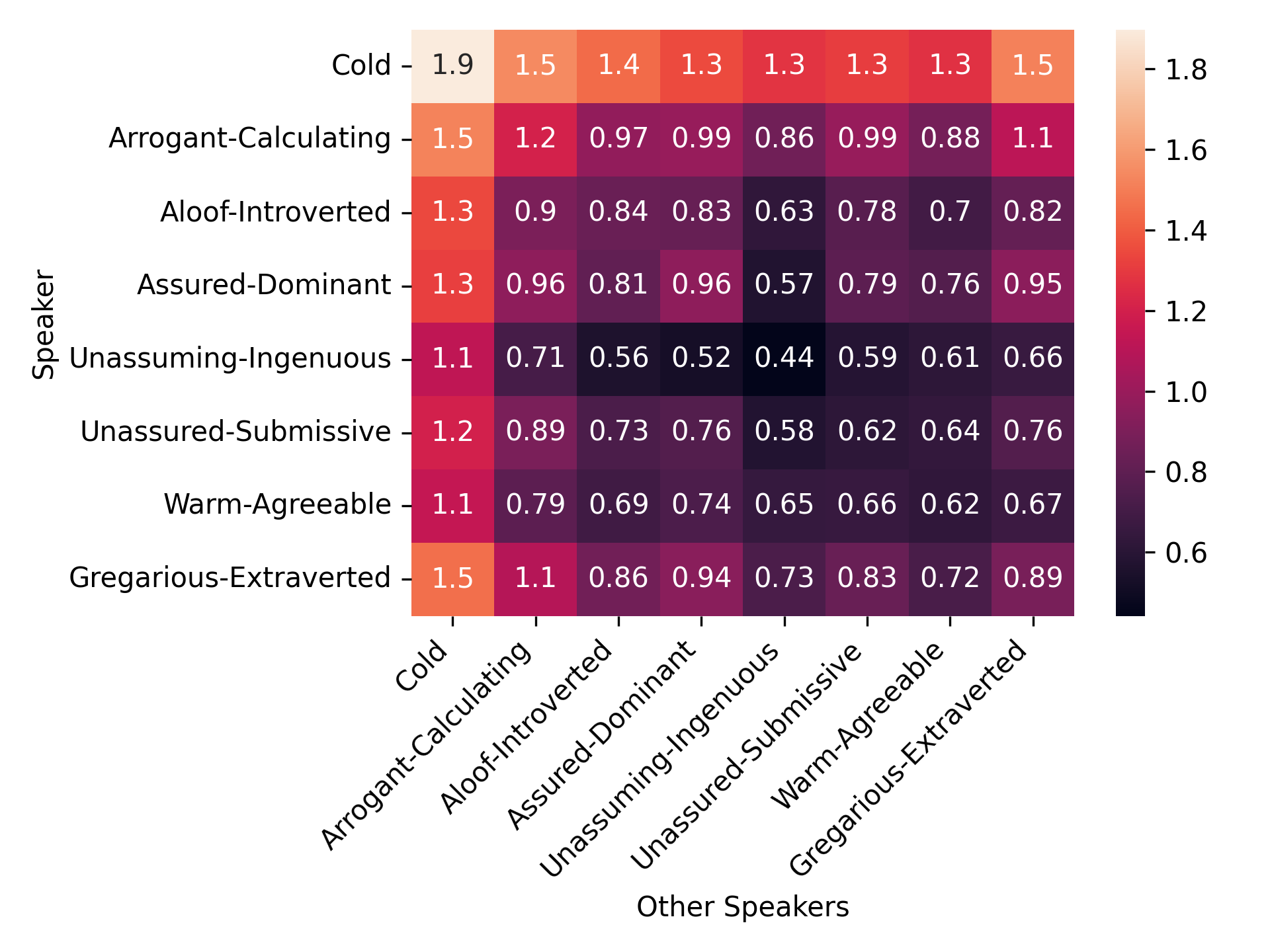}
    \caption{Likelihood ratio of social orientation tags co-occurring in conversations that end in failure to conversations that end in success in the CGA CMV corpus. Ratios greater than 1 mean failure is more likely than success.}
    \label{fig:reddit-social-orientation-co-occurrence}
    \end{center}
\end{figure}

\textbf{Pairs of social orientation tags are more likely co-occur in conversations that succeed or fail in the expected way.} Figure \ref{fig:reddit-social-orientation-co-occurrence} is a co-occurrence matrix for pairs of social orientation tags from the CGA CMV corpus. The matrix was created by first counting the number of occurrences of pairs of social orientation tags. In particular, for each individual speaker we count each of their social orientation tags paired with any of the tags for other speakers. We calculate these counts for all conversations that succeed and again for all conversations that fail. We normalized the resulting 2 matrices so that they form a probability distribution and then we divide the matrix (elementwise) for the failed conversations by the matrix for the successful conversations. The resulting interpretation is that any cell with a value greater than 1 is more likely to occur in a conversation that fails and a value less than 1 is more likely to occur in a conversation that succeeds. We see that \textit{Arrogant-Calculating} and \textit{Cold} comments are more likely to be met by other \textit{Arrogant-Calculating} and \textit{Cold} comments in failed conversations versus successful ones. We see a similar pattern for the successful conversations, where \textit{Unassuming-Ingenuous}, \textit{Unassured-Submissive}, and \textit{Warm-Agreeable} are more likely to be reciprocated (hence the values $<1$). This indicates that certain interaction patterns correlate well with dialogue outcomes in the expected way, which is precisely what the theory attempts to capture.

\section{Conclusion \& Future Work}
\label{sec:conclusion}
In this work, we created a new data set of dialogue utterances labeled with social orientation tags. We demonstrated that social orientation tags outperform text-only models as measured by accuracy for predicting dialogue outcomes in low-resource settings. Further, we showed that even in high-resource settings, the use of social orientation tags improves task performance. Then, we showed how we gain explainability of dialogue outcomes through the use of social orientation features. We concluded with further supporting evidence for the correlation of social orientation features with dialogue outcomes. We release our data sets, code, and models to the public to encourage further research in this area.

Future work can extend the use of social orientation features to other tasks (e.g., negotiations), data sets (e.g., X, formerly Twitter), and languages. We have shown that social orientation features are useful for a Chinese language dialogue outcome prediction task. Our trained \verb|XLMR| model is ready to be used in nearly 100 other languages.

\section*{Limitations}
In our experiments, we found that social orientation tags will perform best in dialogue settings where there is sufficient variation in the social orientation tags. For example, when looking at the Campsite Negotiations Corpus (CaSiNo) \cite{chawla-etal-2021-casino}, we found that most conversation utterances were labeled tags such as \textit{Warm-Agreeable} or \textit{Unassuming-Ingenuous}. This meant that the social orientation tags were not particularly discriminative for predicting dialogue outcomes in this setting. This observations extends to tasks where there is little variation in the task outcome. For example, in the CaSiNo corpus, nearly all conversations participants were satisfied with the outcome of the discussion, meaning there wasn't much of an outcome to predict.

\section*{Ethics Statement}
We see this work as a net-positive tool for being able to examine and moderate uncivil conversations. Furthermore, we see this as a tool for understanding how to improve conversations and training people to navigate difficult social situations (e.g., hiring and firing, etc.). We acknowledge that this tool could be used to more systematically police online speech and that false-positives can stifle discussion. For example, comments that are accidentally flagged as \verb|Cold| can be harmful to open discussion.

\section*{Acknowledgements}
This research is being developed with funding from the Defense Advanced Research Projects Agency (DARPA) Cross-Cultural Understanding program under Contract No HR001122C0034. The views, opinions and/or findings expressed are those of the author and should not be interpreted as representing the official views or policies of the Department of Defense or the U.S. Government.
\newpage
\section*{Appendix}
\subsection{Data Sets}
\begin{table}[!ht]
\begin{center}
\begin{tabular}{rrrrr}
\toprule
Annotator & 1 & 2 & 3 & GPT-4 \\
\midrule
1 & 1.00 & 0.59 & 0.47 & 0.26 \\
2 & 0.59 & 1.00 & 0.59 & 0.20 \\
3 & 0.47 & 0.59 & 1.00 & 0.30 \\
GPT-4 & 0.26 & 0.20 & 0.30 & 1.00 \\
\bottomrule
\end{tabular}
\caption{Inter-annotator agreement rate on the social orientation tagging tasks as measured by the fraction of utterances with agreeing labels.}
\label{tab:inter-annotator-agreement}
\end{center}
\end{table}

\subsection{Definitions of Social Orientation Tags}
We use the following definitions of social orientation tags in this work.

\textbf{Assured-Dominant} - Demands to be the center of interest, demands attention, does most of the talking, speaks loudly, is firm, is self-confident, is forceful, is ambitious, is assertive, is persistent, is domineering, not self-conscious

\textbf{Gregarious-Extraverted} - Feels comfortable around people, starts conversations, talks to a lot of different people, loves large groups, is friendly, is enthusiastic, is warm, is extraverted, is good-natured, is cheerful / happy, is pleasant, is outgoing, is approachable, is not shy, is ``lively''

\textbf{Warm-Agreeable} - is interested in people, reassures others, inquires about others' well-being, gets along well with others, is kind, is polite and courteous, is sympathetic, is respectful, is tender-hearted, is cooperative, is appreciative, is accommodating, is gentle, is charitable

\textbf{Unassuming-Ingenuous} - Tolerates a lot from others, takes things as they come, tells the truth, thinks of others first, does not brag or boast, seldom stretches the truth, does not scheme or plot, is modest, is trustworthy, is unassuming, is honest, not self-centered, is sincere, not demanding, is straightforward

\textbf{Unassured-Submissive} - Speaks softly, lets others finish what they are saying, dislikes being the center of attention, doubts themselves, not especially thorough, doesn't like to work too hard / will give up easily, is impractical, is timid, is inconsistent, is weak, is disorganized, is not authoritative, is a bit lazy, is not forceful

\textbf{Aloof-Introverted} - Is quiet, especially around strangers, is a very private person, doesn't talk a lot / has little to say, doesn't smile much, doesn't reveal much about themselves, is not demonstrative (verbally or non-verbally), is distant, is shy, is impersonal, is introverted, is disinterested in others, is bashful, is not very social, is focused inward

\textbf{Cold} - Believes people should fend for themselves, doesn't fall for sob-stories, is not interested in other people's problems, not warm toward others, is cruel, is ruthless, is cold-hearted, is hard-hearted, is unsympathetic, is uncharitable

\textbf{Arrogant-Calculating} - Flaunts what they have, boasts and brags, will plot and scheme to get ahead, willing to exploit others for own benefit, is big-headed, is tricky, is boisterous, is conniving / calculating, is conceited, is crafty / cunning, is cocky, is manipulative of others

\subsection{Sample Social Orientation Predictions}
Figure \ref{fig:wikipedia_sample_conversation} shows a sample dialogue from the CGA corpus labeled with predicted social orientation tags. The model got most of these tags correct, with the exception of comment (4), which is more likely an \textit{Arrogant-Calculating} comment but the model likely used the winking emoticon to predict \textit{Gregarious-Extraverted}.
\begin{figure}[!ht]
    \begin{minipage}{0.48\textwidth}
        \rule[0.2ex]{\linewidth}{1pt}
        (1) \textbf{Speaker A}: Per wiki\_link. I can and will remove finished discussions/warnings from my talk page. As a sysop, you should be defending my rights not siding with someone in the wrong because you like them better! \textcolor{blue}{\textbf{Assured-Dominant}}
        
        (2) \textbf{Speaker B}: I'm well aware of wiki\_link, and as it says, ``repeated'' replacement of material does nothing but antagonise. However, replacing the contents of your talk page once does not count as repeated, and thus your threat was uncalled for. \textcolor{blue}{\textbf{Assured-Dominant}}
        
        (3) \textbf{Speaker A}: Whatever, i'm still right. \textcolor{blue}{\textbf{Unassured-Submissive}}
        
        (4) \textbf{Speaker B}: Yup, I can clearly see why you started an RfA with an attitude like that. Makes perfect sense ;). \textcolor{blue}{\textbf{Gregarious-Extraverted}}

        (5) \textbf{Speaker A}: Your attitude is terrible for someone who has actually passed an RfA. You seem to pick and chose which policys you want to follow and which to not! I dunno, maybe sysops can! Who knows!?! \textcolor{blue}{\textbf{Cold}}
        \rule[0.2ex]{\linewidth}{1pt}
    \end{minipage}
    \caption{Sample conversation from the CGA data set with social orientation tags are shown in \textcolor{blue}{blue}. This conversation ended in failure due to the last comment.}
    \label{fig:wikipedia_sample_conversation}
\end{figure}

\subsection{Social Orientation Data Set Creation Details}
When we first started the data collection process, we labeled 10 conversations with social orientation tags and fed these conversations to both GPT-3.5-Turbo and GPT-4 and explored several temperature decoding settings and found that $T=0.4$ and GPT-4 performed best with respect to accuracy, which we proceeded with for all remaining API calls. This data set was collected for approximately \$1,000.

We provide definitions of social orientation tags and use 4 labeled conversations in our prompt, which showcase a variety of social orientation tags. We then provide the conversation to be labeled in markdown format. If a conversation is too long to fit into a single prompt, we use an overlap of 1 utterance break the conversation into chunks that fit into a single context window. We generally found that markdown format was most reliable for obtaining structured responses from GPT-4, though occasionally (e.g., 20/4,188), we had to manually fix malformed responses.

Here is the prompt (with an actual input and response) we used with GPT-4 version \verb|gpt-4-0314| to collect social orientation tags for all utterances in the CGA corpus.
\begin{lstlisting}[breakautoindent=false, breakindent=0pt, breaklines]
System Prompt: You are a helpful assistant.

User Input: Social orientation (from circumplex theory) is a social theory that characterizes interactions between speakers. The social orientation tagset includes: {Assured-Dominant, Gregarious-Extraverted, Warm-Agreeable, Unassuming-Ingenuous, Unassured-Submissive, Aloof-Introverted, Cold, Arrogant-Calculating}, which are defined below in more detail.

Assured-Dominant - Demands to be the center of interest, demands attention, does most of the talking, speaks loudly, is firm, is self-confident, is forceful, is ambitious, is assertive, is persistent, is domineering, not self-conscious

Gregarious-Extraverted - Feels comfortable around people, starts conversations, talks to a lot of different people, loves large groups, is friendly, is enthusiastic, is warm, is extraverted, is good-natured, is cheerful / happy, is pleasant, is outgoing, is approachable, is not shy, is \"lively\"

Warm-Agreeable - is interested in people, reassures others, inquires about others' well-being, gets along well with others, is kind, is polite and courteous, is sympathetic, is respectful, is tender-hearted, is cooperative, is appreciative, is accommodating, is gentle, is charitable

Unassuming-Ingenuous - Tolerates a lot from others, takes things as they come, tells the truth, thinks of others first, does not brag or boast, seldom stretches the truth, does not scheme or plot, is modest, is trustworthy, is unassuming, is honest, not self-centered, is sincere, not demanding, is straightforward

Unassured-Submissive - Speaks softly, lets others finish what they are saying, dislikes being the center of attention, doubts themselves, not especially thorough, doesn't like to work too hard / will give up easily, is impractical, is timid, is inconsistent, is weak, is disorganized, is not authoritative, is a bit lazy, is not forceful

Aloof-Introverted - Is quiet, especially around strangers, is a very private person, doesn't talk a lot / has little to say, doesn't smile much, doesn't reveal much about themselves, is not demonstrative (verbally or non-verbally), is distant, is shy, is impersonal, is introverted, is disinterested in others, is bashful, is not very social, is focused inward

Cold - Believes people should fend for themselves, doesn't fall for sob-stories, is not interested in other people's problems, not warm toward others, is cruel, is ruthless, is cold-hearted, is hard-hearted, is unsympathetic, is uncharitable

Arrogant-Calculating - Flaunts what they have, boasts and brags, will plot and scheme to get ahead, willing to exploit others for own benefit, is big-headed, is tricky, is boisterous, is conniving / calculating, is conceited, is crafty / cunning, is cocky, is manipulative of others

In the following conversations drawn from Wikipedia discussion forums, each row corresponds to an Utterance ID, a Speaker ID, and the Text spoken. For each utterance, assign a social orientation tag. Identify the utterance by its Utterance ID and Speaker ID. Here are a few examples.
---
| Utterance ID | Speaker ID | Text |
| --- | --- | --- |
| 1 | Tryptofish | == Good work! == '''The Admin's Barnstar''' For the apparently thankless task of drafting a suggested closing summary at the RfC/U. |
| 2 | The Wordsmith | Thank you for your kindness. I do make an effort to be even-handed, no matter what people wiki_link about me. |
| 3 | Lar | I was just popping by to offer some words of encouragement. Glad to see Tryp beat me to it. ++: / |

| Utterance ID | Speaker ID | Label |
| --- | --- | --- |
| 1 | Tryptofish | Warm-Agreeable |
| 2 | The Wordsmith | Unassuming-Ingenuous |
| 3 | Lar | Warm-Agreeable |
---
| Utterance ID | Speaker ID | Text |
| --- | --- | --- |
| 1 | Gritzko | ==  is under a criminal investigation]] == I am rather pleased to relay that here: Sergey Rublyov aka Ssr who was rather active in curating this article is currently under an actual criminal investigation as his PR services to Mr. Misharin were illegally paid. Namely, 66.ru and politsovet.ru report that Mr Rublyov was provided with a mock employment at a regional energy company as an \"engineer\" (being a journalist by education). The regional prosecutor's office investigates the incident. [EXTERNA_LINK: http://politsovet.ru/40903-delo-o-mertvyh-dushah-rabota-formanchuka-mozhet-zakonchitsya-ugolovnym-delom.html] [EXTERNA_LINK: http://66.ru/news/society/131688/] Regional MP Alshevskikh confirms on Twitter: [EXTERNA_LINK: https://twitter.com/Alshevskix/status/299147903560204289] Well, it was really stupid from Mr Rublyov to do drunk posts on LiveJournal insulting his past employers. |
| 2 | Ssr | Any relation of this info to work on current Wikipedia article? You personally are not recommended to appear here by independent mediators, don't you remember? (because of your and your friends' persistent attempts to violently use Wikipedia for political attacks\u2014while I was acting correctly according to rules, see also Russian article/talk\u2014and your edits to both were totally wiped out) No \"past employers\" were insulted BTW.  |
| 3 | 2A02:6B8:0:107:D83D:EE04:EA8D:1553 | How unfortunate, I am not illegally employed full-time to whitewash reputations of corrupted politicians. Hence, I do not have that much time to defend my edits. But maybe, I will make another attempt. |
| 4 | Ssr | I doubt you are able (if this long number is you) because _several_ independent mediators in ru end en after many long-time investigations decided that you and your friends try to violate wikipedia for political purposes so no luck for you here (read posts above including links to mediations\u2014don't forget!). Such posts as this particular your post are not welcome here because it's unrelated to work on the article and may be deleted as off-topic (in ru this is widely practiced). |
| 5 | Gritzko | I am pretty sure you are not talking to me now because you certainly know that I certainly know that you are lying. I'll make my edits this Saturday, be prepared. |
| 6 | Ssr | No need for me to be prepared, I am, as you, a COI party, am not going to edit, and mediators will do their job well as they did before (many thanks again to them for their great work). |
| 7 | Ssr | *Also, there's an arbitraiton warning for other Gritzko's violations: wiki_link, so he must be under strict control, as his \"warnings\" most probably indicate further violations. |
| 8 | Gritzko | You have a COI cause you were paid to doctor this article. I have no COI. You are a liar. Clear enough? |

| Utterance ID | Speaker ID | Label |
| --- | --- | --- |
| 1 | Gritzko | Cold |
| 2 | Ssr | Unassuming-Ingenuous |
| 3 | 2A02:6B8:0:107:D83D:EE04:EA8D:1553 | Arrogant-Calculating |
| 4 | Ssr | Warm-Agreeable |
| 5 | Gritzko | Arrogant-Calculating |
| 6 | Ssr | Unassuming-Ingenuous |
| 7 | Ssr | Unassuming-Ingenuous |
| 8 | Gritzko | Cold |
---
| Utterance ID | Speaker ID | Text |
| --- | --- | --- |
| 1 | DarkHero | check the sig. Leaked Info? |
| 2 | Sukecchi | I highly doubt those are real. - I doubt it too...but I still wonder |
| 3 | BrydoF1989 | Fat chance. These appear to be the creatures from Telefang O_o |
| 4 | Joizashmo | I think they look more like Digimon than Pok\u00e9mon. |
| 5 | Rat235478683 | The middle one looks like the evolved form of Heracross.Rat235478683 |
| 6 | 68.65.113.160 | Hi, I'm Kojiro who had that sig. They were indeed Telefang, I just wanted too see how many people believed it. XD I didn't mean to cause any trouble. |
| 7 | Rat235478683 | You stink!Rat235478683 |

| Utterance ID | Speaker ID | Label |
| --- | --- | --- |
| 1 | DarkHero | Aloof-Introverted |
| 2 | Sukecchi | Unassured-Submissive |
| 3 | BrydoF1989 | Arrogant-Calculating |
| 4 | Joizashmo | Unassuming-Ingenuous |
| 5 | Rat235478683 | Unassuming-Ingenuous |
| 6 | 68.65.113.160 | Unassured-Submissive |
| 7 | Rat235478683 | Cold |
---
| Utterance ID | Speaker ID | Text |
| --- | --- | --- |
| 1 | Jack1234567891011121314151617 | Multiculturalism So i think it's false to say that the alt right opposes multiculturalism. Because from what i've understood they basically want an white ethno state where all europeans would be welcome. How would they have an white ethno state without many cultures? The alt right might say they are against it but they don't seem to  understand that they basically advocate for a multicultural state |
| 2 | Lukacris | While race and culture aren't exactly the same thing, I think it's fair to conceptualize white nationalism as antithetical to multiculturalism. |
| 3 | Jack1234567891011121314151617 | @Lukacris I know that race and culture aren't the same thing obviously, but what i meant is that ultimate goal for the alt right is the ethno state where ALL whites would be welcome. If all whites with many different cultures  are welcome then they're for multiculturalism.  \u2014\u00a0Preceding unsigned comment added by |
| 4 | Beyond My Ken | This is general discussion of the topic  i.e. WP:NOTAFORUM  and not about how to improve the article.  If Jack123... has a reliable source that says that the alt-right is not opposed to mutliculturalism. he should provide it. |
| 5 | Jack1234567891011121314151617 | @Beyond My Ken Don't you think that the alt right advocating for an ethno state is evidence enough? |
| 6 | Beyond My Ken | Nope, nowhere near enough. In fact, your stretching of their position strains credulity, since it's completely bullshit. |

| Utterance ID | Speaker ID | Label |
| --- | --- | --- |
| 1 | Jack1234567891011121314151617 | Gregarious-Extraverted |
| 2 | Lukacris | Warm-Agreeable |
| 3 | Jack1234567891011121314151617 | Arrogant-Calculating |
| 4 | Beyond My Ken | Unassuming-Ingenuous |
| 5 | Jack1234567891011121314151617 | Unassured-Submissive |
| 6 | Beyond My Ken | Cold |
---
| Utterance ID | Speaker ID | Label |
| --- | --- | --- |
| 1 | Hipocrite | This kind of stuff dosen't get sent around enough, so the bad seems to build up and overpower the good. I have always had a great deal of respect for your decision making abilities. The case you tried to solve was a difficult and contentious one, and I fully support you walking away from it without reaching strong conclusion. You are a bigger man for doing so. I tried to figure out who was right, but also got fully frusterated by the intracasies of what must have been going on for years. Please don't think less of yourself or let the slings and arrows of whomever is shooting at you hit. You are a good and valuable contributor and problem solver.  - |
| 2 | Hiding | *Look, I really appreciate your words, but right this second, and believe me, this isn't easy to write, it isn't enough.  I've been building all this up for ages, because we have this good faith idea and we have this idea of being civil, and it makes it so hard to vent. I really don't think the community values problem solvers anymore, because every single decision made on Wikipedia is now open to review on so many counts it has become an institutionalised bureaucracy.  Consideration is seriously undervalued on Wikipedia and I don't really see a way of improving it.  I know in my own head I haven't got the make up to go rouge, but I also know I can't be chipped at forever.  I'm fed up to my high teeth of all the back biting. |
| 3 | Hipocrite | ** The \"community\" is shit. The part of the community that you care about - the people who want to empower other people to write an encyclopedia, values problem solvers. We really do - I promise. Decisions on wikipedia are reviewed by people who believe it is an institutionalised bureaucracy, and then the decisions are actually made the way they should be - by people who get it. I *know* it dosen't feel that way for you *right now* - and it didn't feel that way for me external_link, but, and not to put too agressive of a spin on it - we are/were wrong. Who is biting your back? Some putrid little troll? Some ultra-inclusionist Wikipedia Review kibitzer? An Encyclopedia Dramatica internet whore? I know it's impossible to ignore them (I've tried and failed), but it's not hard to pretend to ignore them, and it's really really satisfying. Create a legitmate sockpuppet and start fixing random articles, or just walk away. While we need you, we don't need you right now, and you don't need us. Do you play Chess, World of Warcraft or Battlefield 2142? I'll let you win or lose at any of them - your call.  - |

Assistant Response:
| Utterance ID | Speaker ID | Label |
| --- | --- | --- |
| 1 | Hipocrite | Warm-Agreeable |
| 2 | Hiding | Unassured-Submissive |
| 3 | Hipocrite | Warm-Agreeable |
\end{lstlisting}

\subsection{Chinese Dialogue Outcome Prediction Data Set Creation Details}

We utilized all available Chinese data from the WikiConv data set \citep{hua-etal-2018-wikiconv}, which comprises 7,731,744 utterances. Our criteria for conversation selection included having at least two unique speakers and a minimum of five conversational turns, resulting in 31,048 conversations with a total of 274,198 utterances. We processed all utterances through COLDETECTOR \citep{deng-etal-2022-cold}, identifying 3,473 potentially toxic comments, approximately 1.3\% of the total.

We then applied two additional filters to the conversations: the presence of at least two toxic utterances and a minimum of two conversational turns involving distinct speakers before the first toxic comment. This narrowed the data set to 360 conversations. Subsequently, two annotators evaluated the first two toxic comments in each conversation for toxicity. If both comments were deemed non-toxic by at least one annotator, the conversation was excluded, leaving 267 conversations. These were then randomly paired with another conversation from the same Wikipedia Talk page. Unmatched conversations were also discarded, culminating in a final data set of 234 conversation pairs, or 468 conversations in total.

\label{cn_dataset_appendix}
\subsection{Social Orientation Tagger Model Details}
We initialize our social orientation tagger weights from the \verb|distilbert-base-uncased| pre-trained checkpoint from Hugging Face\footnote{\url{https://huggingface.co/distilbert-base-uncased}}. We use following hyperparameter settings: batch size=32, learning rate=1e-6, we include speaker names before each utterance, we train in 16 bit floating point representation, we use window size of 2 utterances (i.e., we use the previous utterance's text and the current utterance's text to predict the current utterance's social orientation tag, and we use a weighted loss function to address class imbalance and improve prediction set diversity. The weight $w'_c$ assigned to each class $c$ is defined by

\begin{align*}
    w'_c = C \cdot \frac{w_c}{\sum_{c=1}^C w_c}
\end{align*}

where $w_c = \frac{N}{N_c}$, where $N$ denotes the number of examples in the training set, and $N_c$ denotes the number of examples in class $c$ in the training set, and $C$ is the number of classes. In our case is $C=9$, including the \textit{Not Available} class, which is used for all empty utterances.

\subsection{Dialogue Outcome Prediction Model Details}
We initialize our dialogue outcome prediction model's weights from the \verb|distilbert-base-uncased| pre-trained checkpoint from Hugging Face\footnote{\url{https://huggingface.co/distilbert-base-uncased}}. We use following hyperparameter settings: batch size=32, learning rate=5e-6, we include speaker names before each utterance, we train in 16 bit floating point representation, and when we include social orientation features, they are included after the speaker's name but before the speaker's utterance. We use all available utterances in a dialogue, with the exception of the last turn in CGA, which is the utterance that signifies a conversation has succeeded or failed. We also include all social orientation tags in the vocabulary of the model so that it can learn custom embeddings for these tags. We train the model 5 times at each training data set size in the set $\{0.01, 0.10, 0.2, 0.5, 1.0\}$ and vary the random seed in the range 42 through 46, which results in 25 model training runs per experimental setting. We train all models on a single instance with 8 NVIDIA RTX A6000 GPUs.

\newpage
\nocite{*}
\section{Bibliographical References}\label{sec:reference}

\bibliographystyle{lrec-coling2024-natbib}
\bibliography{lrec-coling2024}

\end{document}